\documentclass[conference]{IEEEtran}
\IEEEoverridecommandlockouts
\usepackage{times}
\usepackage{soul}
\usepackage{setspace}
\usepackage{url}
\usepackage[hidelinks]{hyperref}
\usepackage[utf8]{inputenc}
\usepackage[small]{caption}
\usepackage{subcaption}
\usepackage{multirow}
\usepackage{graphicx}
\usepackage{amsmath}
\usepackage{flushend}
\usepackage{amsmath}
\usepackage{comment}
\usepackage{amsthm}
\usepackage{booktabs}
\usepackage{algorithmic}
\usepackage{balance}

\usepackage[ruled,noend,linesnumbered]{algorithm2e} 

\def\BibTeX{{\rm B\kern-.05em{\sc i\kern-.025em b}\kern-.08em
		T\kern-.1667em\lower.7ex\hbox{E}\kern-.125emX}}
\begin{document}
	
	\title{HexCNN: A Framework for Native Hexagonal Convolutional Neural Networks
	}
	
	\author{%
		\IEEEauthorblockN{Yunxiang Zhao\IEEEauthorrefmark{2}, Qiuhong Ke\IEEEauthorrefmark{2}, Flip Korn\IEEEauthorrefmark{3}, Jianzhong Qi\IEEEauthorrefmark{2}, Rui Zhang\thanks{\IEEEauthorrefmark{1}Rui Zhang is the corresponding author} \IEEEauthorrefmark{2}\IEEEauthorrefmark{1}}
		\IEEEauthorblockA{\IEEEauthorrefmark{2}The University of Melbourne, Australia,
			\IEEEauthorrefmark{3}Google Research, USA\\
			\{yunxiangz@student., qiuhong.ke@, jianzhong.qi@, rui.zhang@ \}unimelb.edu.au, flip@google.com} 
	}
	
	\maketitle
	
	\begin{abstract}
		Hexagonal CNN models have shown superior performance in applications such as IACT data analysis and aerial scene classification due to their better rotation symmetry and reduced anisotropy. 
		In order to realize hexagonal processing, existing studies mainly use the ZeroOut method to imitate hexagonal processing, which causes substantial memory and computation overheads.
		We address this deficiency with a novel native hexagonal CNN framework named \emph{HexCNN}. HexCNN takes hexagon-shaped input and performs forward and backward propagation on the original form of the input based on hexagon-shaped filters, hence avoiding computation and memory overheads caused by imitation. 
		For applications with rectangle-shaped input but require hexagonal processing, HexCNN can be applied by padding the input into hexagon-shape as preprocessing. In this case, we show that the time and space efficiency of HexCNN still outperforms existing hexagonal CNN methods substantially.
		Experimental results show that compared with the state-of-the-art models, which imitate hexagonal processing but using rectangle-shaped filters, HexCNN reduces the training time by up to 42.2\%.
		Meanwhile, HexCNN saves the memory space cost by up to 25\% and 41.7\% for loading the input and performing convolution, respectively. 
		
	\end{abstract}
	
	
	\section{Introduction}
	\label{sec:introduction}
	Recent studies show that compared with traditional rectangle-based CNN models, CNN models with hexagon-shaped filters achieve better performance in applications such as Imaging Atmospheric Cherenkov Telescope (IACT) data analysis~\cite{erdmann2018deep,moriakov2020inferring,nieto2019studying,shilon2019application}, Hex move-prediction~\cite{young2016neurohex}, and IceCube data analysis~\cite{huennefeld2017deep}. Applying hexagonal filters in group CNNs can even surpass the performance of traditional CNN models with image classification tasks on data sets such as CIFAR-10 ~\cite{hoogeboom2018hexaconv,schlosser2019hexagonal,sun2016design}.
	
\begin{figure}[tb]
		\centering
		\setlength{\belowcaptionskip}{-0.5cm}   
		\includegraphics[width=1\columnwidth]{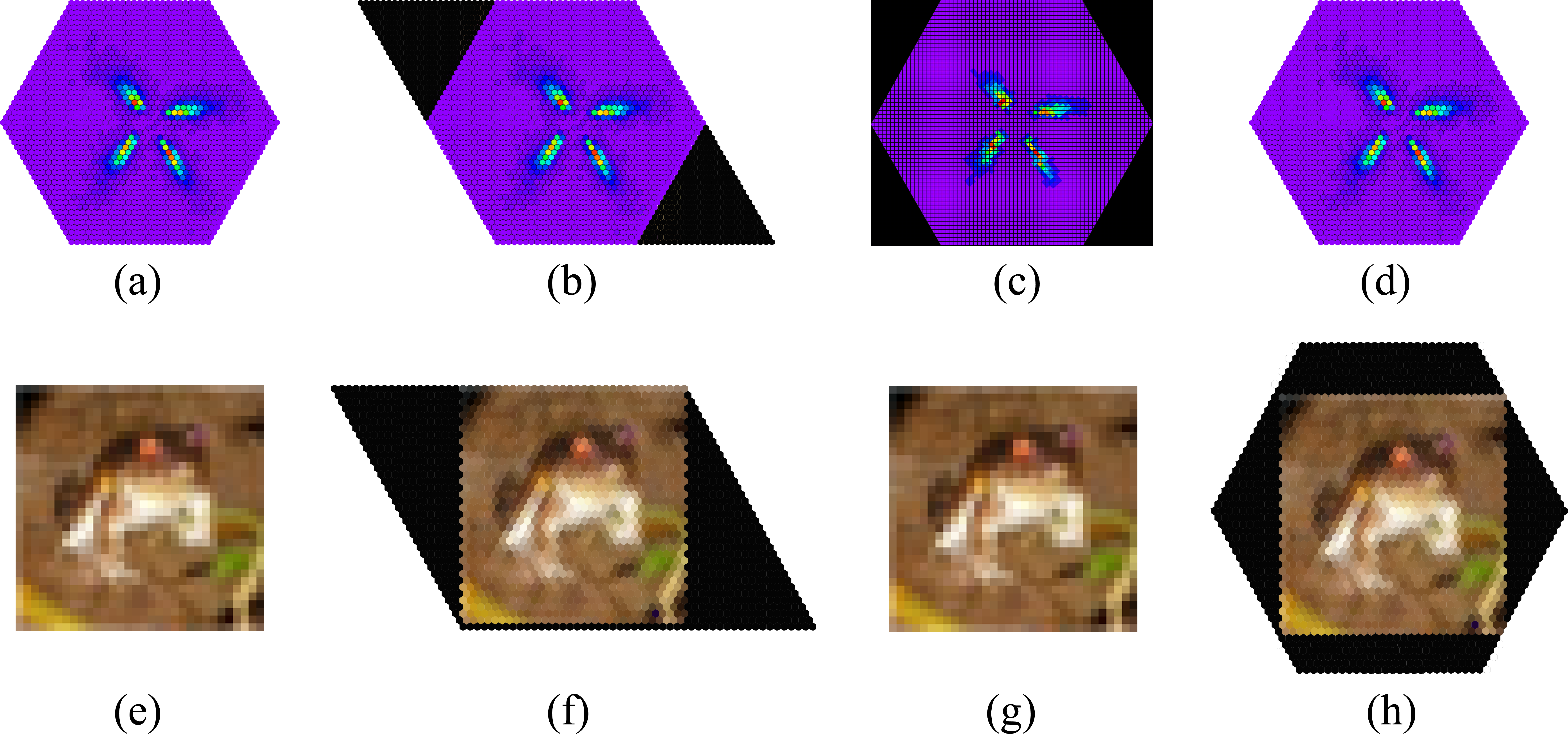}
		\caption{The original inputs (a, e), and the results after pre-processing using ZeroOut \protect\cite{hoogeboom2018hexaconv} (b, f), Quasi-H~\cite{sun2016design} (c, g), and HexCNN (d, h). (a) a hexagonal telescope image of gamma-ray events \protect\cite{mangano2018extracting}. (e) a rectangle-shaped image from CIFAR-10 \protect\cite{krizhevsky2009learning} (best view in color).}
		\label{fig:1}
	\end{figure}

	To realize hexagonal processing, most existing studies apply rectangle-shaped filters with the ZeroOut method to imitate hexagonal processing~\cite{hoogeboom2018hexaconv,luo2019hexagonal,ke2018hexagon}. 
	We refer to these models as hexagon-imitation models. 
	These models, however, require a padding strategy as a pre-processing for the input. Fig.~\ref{fig:1}b and Fig.~\ref{fig:1}c illustrate the idea of such padding on the hexagon and rectangle-shaped input, and resampling is required if the input is not hexagonal grids such as Fig~\ref{fig:1}e. The padded area (dark elements in Fig.~\ref{fig:1}b and Fig.~\ref{fig:1}c) is not needed for the output but is computed as part of the input throughout the hexagon-imitation models. This is due to the restrictions of existing deep learning frameworks where data is represented in a rectangular way.
	After padding the input, the rectangle-shaped filters in the network can be used to process the input without having to accommodate the boundary of hexagonal grids.
	To achieve hexagonal processing in convolution and pooling, specific positions of the rectangle-shaped filter are set to zero to eliminate their influence on the output (e.g., the two blue-colored elements with ``0" in Fig.~\ref{fig:2}a). These zeroed-out elements, however, lead to unnecessary memory and computation costs in hexagon-imitation models.
	To sum up, the limitation of hexagon-imitation models is that the padding strategy and the ZeroOut manner in convolution and pooling cause significant memory and computation overhead for both hexagon and rectangle-shaped input.
	
	In this paper, we introduce a new framework called HexCNN to address this limitation. HexCNN takes hexagon-shaped input and performs hexagonal forward and backward propagation on the original form of the input based on hexagon-shaped filters. We refer to this method as ``native hexagonal processing". Compared with previous methods, the proposed native hexagonal processing eliminates the memory and computation overhead from the padding and ZeroOut operations. 
	As shown in Fig.~\ref{fig:1}d, for hexagon-shaped input, the proposed HexCNN does not perform padding as done in hexagon-imitation models. 
	The convolution is achieved using hexagon-shaped filters, which bypasses the unnecessary computation of the zeroed-out elements, as shown in Fig.~\ref{fig:2}b. 
	Therefore, HexCNN saves memory and computation costs significantly when loading the input and performing filter related operations such as convolution, pooling, and backpropagation. For rectangle-shaped input such as images, HexCNN incurs resampling and padding, and the incurred unnecessary area has a similar size to those in hexagon-imitation models, as shown in Fig.~\ref{fig:1}f, and Fig.~\ref{fig:1}h. In this case, HexCNN still has less overhead due to the reduced space and computation cost on native hexagonal convolution. We detail the comparison between HexCNN and hexagon-imitation models on both hexagon and rectangle-shaped input in Section~\ref{sec:experiment}.
	We summarize our contributions as follows:
	
	\begin{figure}[t]
		\centering
		\setlength{\belowcaptionskip}{-0.5cm}   
		\begin{subfigure}[a]{0.19\linewidth}
			\includegraphics[width=\linewidth]{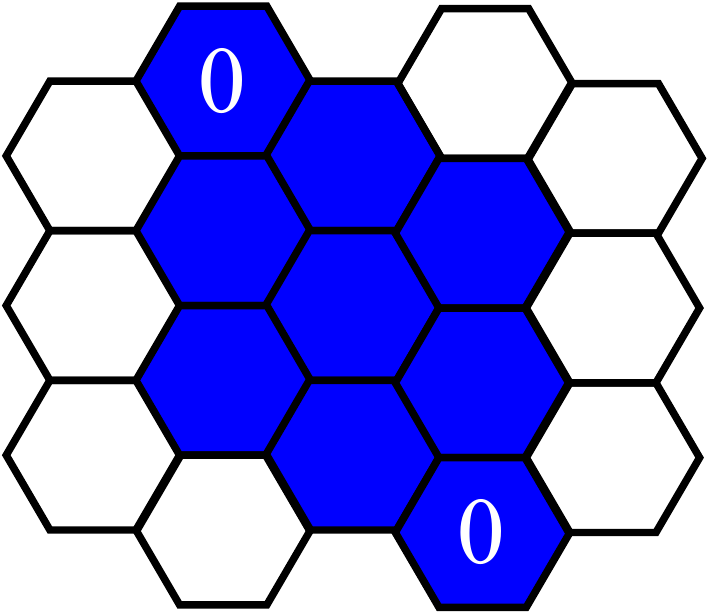}
			\caption{}
		\end{subfigure}\,\,\,
		\begin{subfigure}[a]{0.19\linewidth}
			\includegraphics[width=\linewidth]{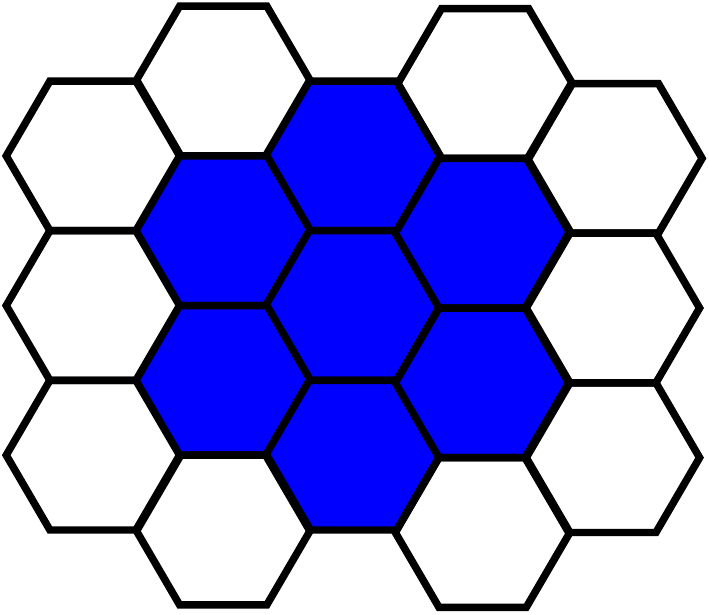}
			\caption{}
		\end{subfigure}\,\,\,
		\begin{subfigure}[a]{0.215\linewidth}
			\includegraphics[width=\linewidth]{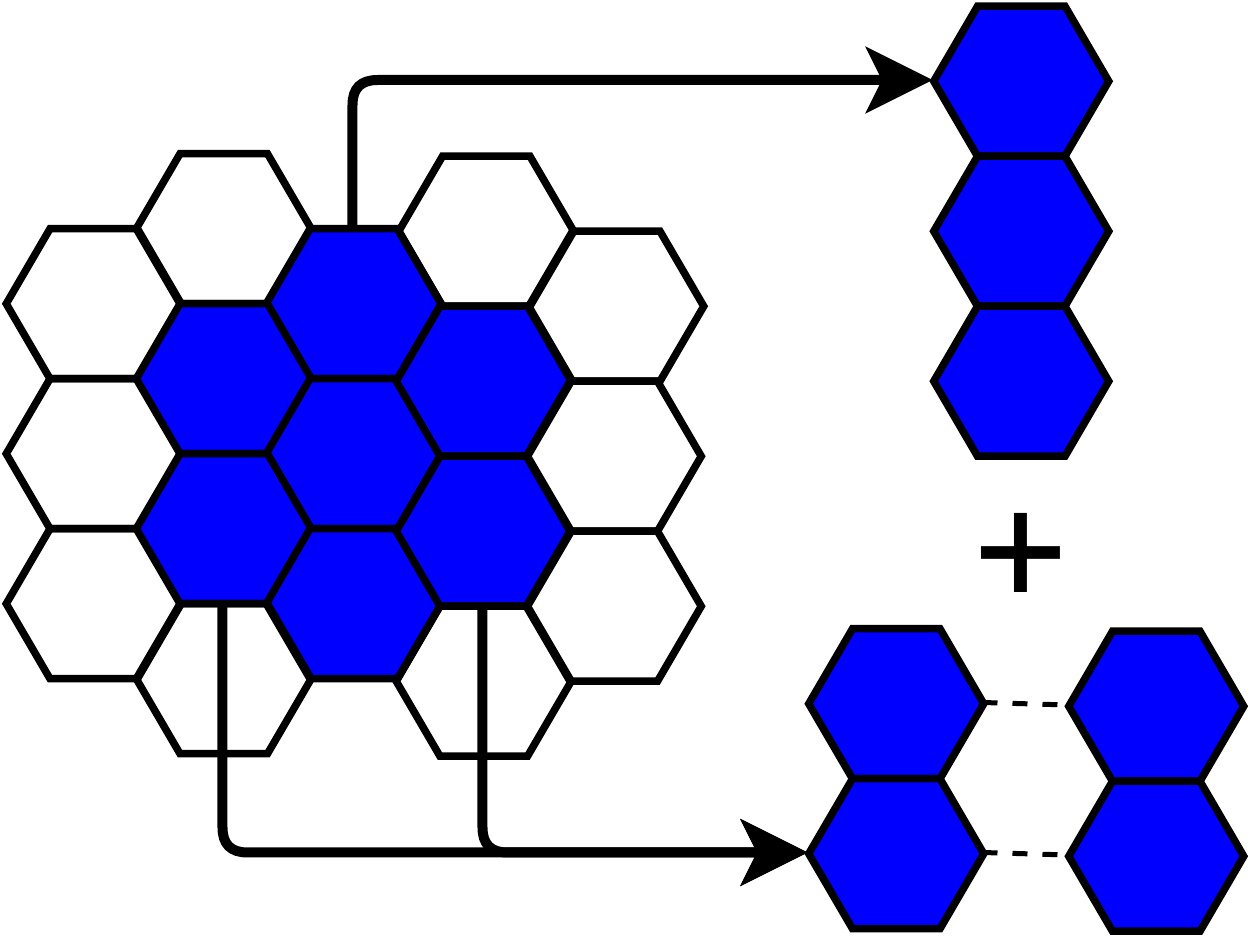}
			\caption{}
		\end{subfigure}\,\,\,
		\begin{subfigure}[a]{0.28\linewidth}\,
			\includegraphics[width=\linewidth]{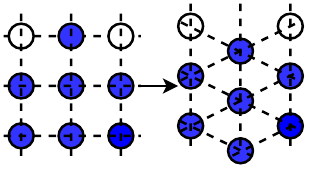}
			\caption{}
		\end{subfigure}
		\caption{Hexagonal convolution via (a) ZeroOut \protect\cite{hoogeboom2018hexaconv}, (b) HexCNN, (c) HexagDLy \protect\cite{steppa2019hexagdly}, and (d) Quasi-H \protect\cite{sun2016design}. Elements in blue denote the filter (best view in color).}
		\label{fig:2}
	\end{figure}
	
	\begin{itemize}
		\item We propose a framework for realizing native hexagonal processing named HexCNN. HexCNN provides native hexagonal CNN operation algorithms for both forward and backward propagation to achieve superior time and space efficiency.
		
		\item We propose an algorithm for transforming hexagonal convolution into matrix multiplication so as to exploit matrix optimization algorithms for reducing the time consumption of hexagonal convolution.
		
		\item We implement HexCNN on TensorFlow and perform extensive experiments to evaluate its efficiency. The experimental results show that 
		HexCNN outperforms the state-of-the-art models, which imitate hexagonal processing, by reducing up to 42.2\% of the training time.
		Meanwhile, HexCNN saves the memory cost by up to 25\% and 41.7\% for loading the input and performing convolution, respectively. 
		For applications with rectangle-shaped input but require hexagonal processing, HexCNN still outperforms existing hexagonal CNN methods by reducing more than 20\% of the training time.
	\end{itemize}
	
	We organize the rest of this paper as follows. We review related work in Section~\ref{sec:relatedWork} and present the fundamentals of HexCNN in Section~\ref{sec:hexCNN}. We describe the algorithm for transforming hexagonal convolution to matrix multiplication in Section~\ref{sec:ctom}, and report the experimental results in Section~\ref{sec:experiment}. We conclude the paper in Section~\ref{sec:conclusion}.
	
	\section{Related Work}
	\label{sec:relatedWork}
	In this section, we review studies that imitate hexagonal processing. Existing hexagon-imitation models only work on rectangle-shaped input. The rectangle-shaped input can be hexagonal grids or square grids, as shown in Fig.~\ref{fig:2}a and Fig.~\ref{fig:2}d.
	We classify existing hexagon-imitation models into two categories. The first category is for input that is rectangle-shaped hexagonal grids (Section~\ref{hexForspa}) and the second category is for input that is rectangle-shaped square grids (Section~\ref{hexForimg}).
	
	\subsection{Hexagonal CNN Models on Hexagonal Grids}
	\label{hexForspa}
	In order to realize hexagonal processing on hexagonal grids, the ZeroOut method has been widely used~\cite{hoogeboom2018hexaconv,luo2019hexagonal}. Young et al.~\cite{young2016neurohex} propose a hexagon-imitation model for a chess game Hex and apply ZeroOut to achieve hexagonal processing. 
	Ke et al.~\cite{ke2018hexagon} propose a hexagon-imitation model for forecasting the ride-sourcing supply and demand, which splits the input into a hexagonal lattice and then realizes hexagonal processing by applying the rectangle-shaped filters together with transform matrices. The concept of transform matrices is similar to that of ZeroOut.
	A similar method was proposed by Jacquemont~\cite{jacquemont2019indexed}, where the values of specific positions in the filter are eliminated by a mask matrix. For convolution, the models above mainly use rectangle-shaped filters together with ZeroOut methods to achieve hexagonal processing~\cite{kerr2012novel}. We use \textbf{ZeroOut} to represent above models.
	
	Steppa and Holch~\cite{steppa2019hexagdly} propose \textbf{HexagDLy}, which implements hexagonal convolution and pooling by combining multiple rectangle-shaped filters with different dilations and sizes, as shown in Fig.~\ref{fig:2}c.
	HexagDLy does not use the ZeroOut method during the convolution and achieves better efficiency along with the increase of filter size due to the advantage of divide-and-conquer. 
	However, HexagDLy only supports convolution and pooling.
	
	\subsection{Hexagonal CNN Models on Square Grids}
	\label{hexForimg}
	To perform hexagonal CNNs on square grids such as images and videos~\cite{hoogeboom2018hexaconv,kerr2012novel,sun2016design}, it is required to first transform the input from square grids to hexagonal grids. Resampling and zero-padding have been widely used for this transformation, which introduces unnecessary computation and space costs, as illustrated in Fig.~\ref{fig:1}f.
	For convolution, existing work mainly uses rectangle-shaped filters together with the ZeroOut method to achieve hexagonal processing~\cite{kerr2012novel}, as shown in Fig.~\ref{fig:2}a. We consider them as \textbf{ZeroOut}, which is the same as that in Section~\ref{hexForspa}. The only difference is that the ZeroOut here takes square grids as the input, and the one in Section~\ref{hexForspa} takes hexagonal grids. 
	
	Sun et al.~\cite{sun2016design} propose quasi-hexagonal kernels (\textbf{Quasi-H}) for different convolution layers.
	Quasi-H is not hexagonal processing in the strict sense and can only be considered as hexagonal processing on deep CNN models with multiple convolutional layers (each layer has multiple filters).
	Applying Quasi-H on rectangle-shaped input for simulating hexagonal convolution does not incur resampling or unnecessary padding, as shown in Fig~\ref{fig:1}g. 
	Applying Quasi-H to hexagon-shaped input requires resampling and padding for pre-processing, which the same as ZeroOut, as shown in Fig.~\ref{fig:1}c. 
	
	\section{HexCNN}
	\label{sec:hexCNN}
	Existing deep learning frameworks such as TensorFlow, PyTorch and Caffe are efficient and maintained by open source developers to keep optimizing their performance~\cite{abadi2016tensorflow,jia2014caffe,paszke2017automatic,zhao2019cbhe}. 
	Instead of implementing HexCNN from scratch, we have implemented it based on the TensorFlow by adapting the core libraries and third-party Eigen library of TensorFlow, so that HexCNN retains the general applicability of the framework.
	This way, implementing hexagonal CNN models of various neural network structures on HexCNN is similar to implementing rectangular CNN models of various structures on TensorFlow. Therefore, HexCNN is a framework for implementing hexagonal CNN models, i.e., CNN models that directly take hexagonal inputs.
	The tensor in TensorFlow is rectangle-shaped, and all the tensor-related steps need to be rewritten for hexagonal processing. In TensorFlow, the input is rectangular. For example, an RGB image is represented as a 3D tensor \textbf{[height, width, channel]}, while in HexCNN, the input is hexagonal, and the corresponding tensor is represented as \textbf{[side\_length, channel]}.
	In this section, we first give an overview of HexCNN followed by its data representation and fundamental operations.
	
	\subsection{Overview}
	\label{hexCNNCoor}
\begin{figure}[tb]
		\centering
		\setlength{\belowcaptionskip}{-0.5cm}   
		\includegraphics[width=0.8\columnwidth]{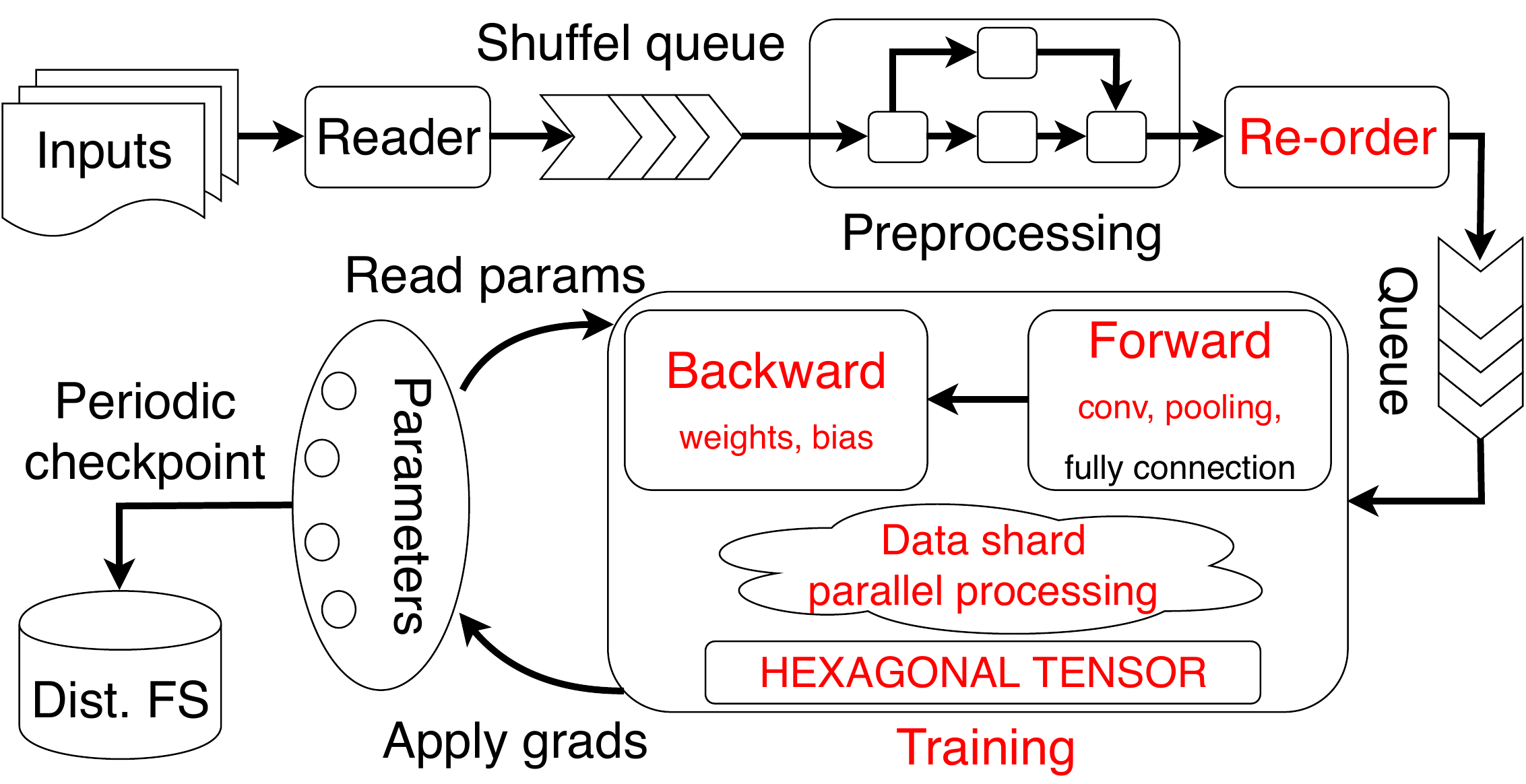}
		\caption{The HexCNN dataflow graph inherited from TensorFlow \protect\cite{abadi2016tensorflow}. HexCNN achieves native hexagonal processing for forward and backward propagation, and the low-level parallel and data sharding algorithms, which are marked in red (best view in color).}
		\label{fig:3}
	\end{figure}

	Fig.~\ref{fig:3} is the schematic dataflow graph of HexCNN, the ``Re-order" operation feeds an input into a hexagon-shaped tensor.
	For optimization purposes, TensorFlow performs parallel computing and data sharding, which are highly related to the shape of the input, filters, and how they are stored in the memory. HexCNN adapts these algorithms according to the hexagonal data representation, which is detailed in Section~\ref{sec:hexCNNds}.
	HexCNN applies hexagon-shaped filters for convolution and pooling during forward-propagation. For backpropagation, HexCNN converts the gradient computation into convolution, and hence it is also based on hexagonal convolution. We detail the hexagonal operations in Section~\ref{sec:conv}. 
	
	\subsection{Data Representation and Storage}
	\label{sec:hexCNNds}
	
\begin{figure}[tb]
		\centering\setlength{\belowcaptionskip}{-0.5cm}   
		\begin{subfigure}[a]{0.4\linewidth}
			\includegraphics[width=\linewidth]{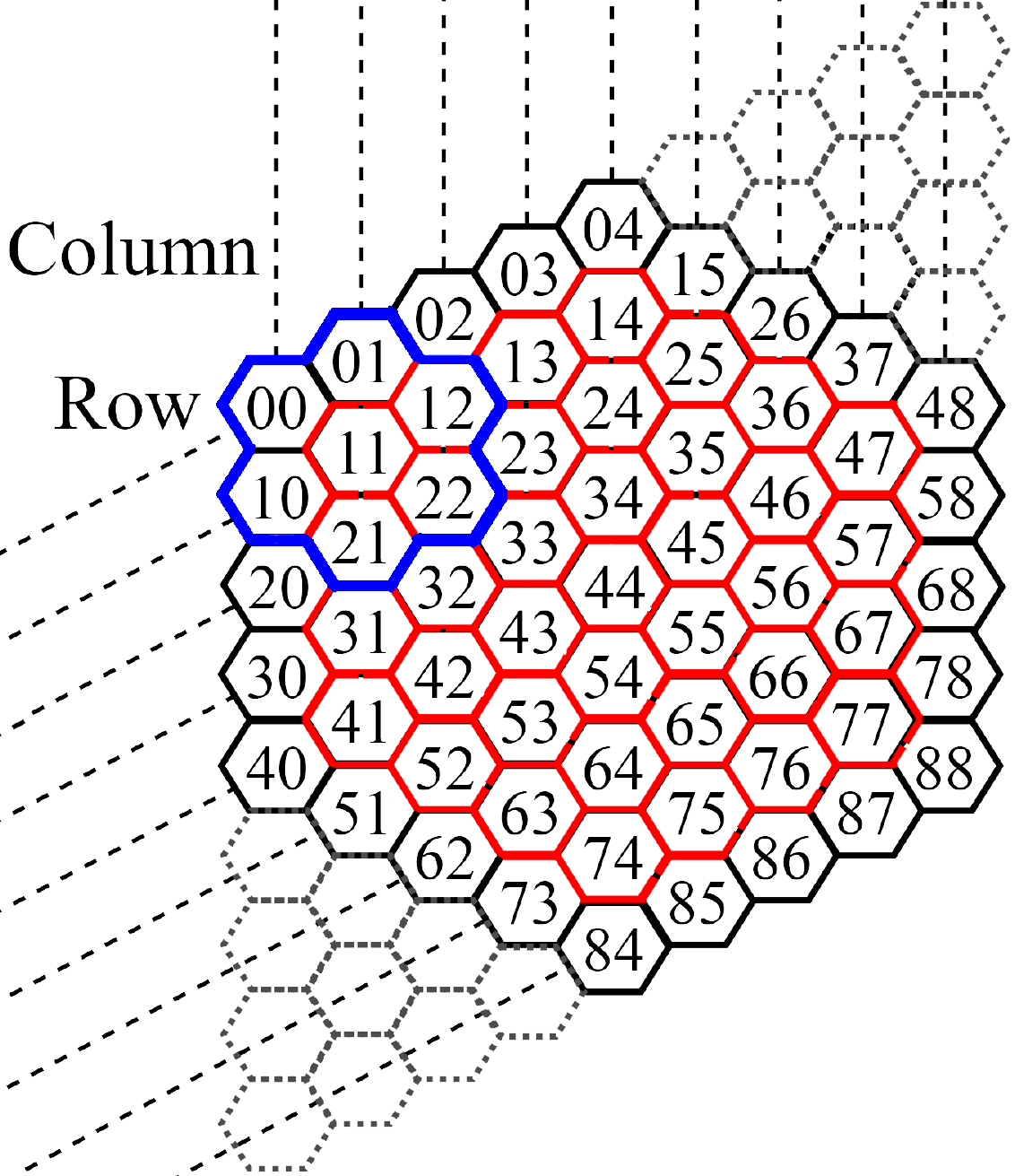}
			\caption{Data representation}
		\end{subfigure}\,
		\begin{subfigure}[a]{0.4\linewidth}
			\includegraphics[width=\linewidth]{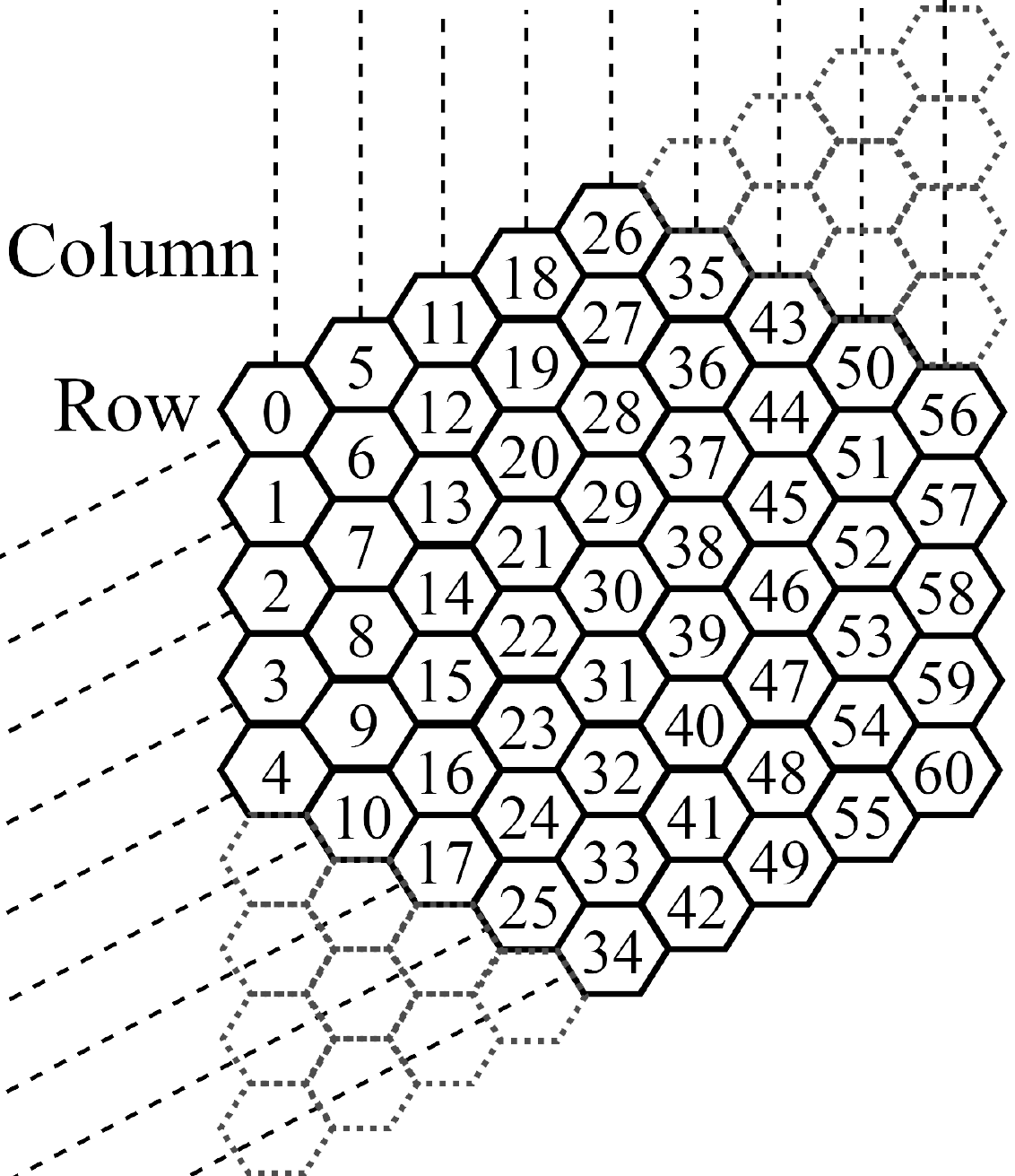}
			\caption{Data storage}
		\end{subfigure}
		\caption{Data representation (a) and storage (b) of HexCNN. HexCNN is based on 2D Axial coordinate system, and transform the input into a column major vector for data storage (best view in color).}
		\label{fig:4}
	\end{figure}

	One of the fundamentals of HexCNN is data representation, which determines the way to perform hexagonal processing. 
	HexCNN takes the idea of the Axial coordinate system~\cite{her1995geometric,hoogeboom2018hexaconv}, and moves the origin to the top-left corner, as shown in Fig.~\ref{fig:4}a.
	Suppose the side length of the hexagon-shaped input is $k$. Then the length of the first $k$ rows increases by one for each row and reduces by one after the $k^{th}$ row. 
	
	To reduce the memory and computation costs of hexagonal processing, HexCNN transforms the hexagon-shaped input into a column-based vector during the data storage phase. Compared with hexagon-imitation models, HexCNN does not require unnecessary spaces for storing both the input and convolution filters. Thus, it has lower memory and computation costs. The dashed elements in Fig.~\ref{fig:4} illustrates the space that HexCNN saves for storing the input and filters in comparison to existing hexagon-imitation models. 

	In convolution, existing hexagon-imitation models mainly use ZeroOut to realize hexagonal convolution. This approach, requires extra space for packing the filter, so as to run on existing deep learning frameworks, as shown in Fig.~\ref{fig:2}a.
	In pooling, both HexCNN and hexagon-imitation models iteratively go through the input, and they have the same space complexity. However, extra computation is required by hexagon-imitation models due to the padding strategy applied to the input.
	
	\subsection{Foundamental Operations}
	\label{sec:conv}
	For rectangle-shaped tensors, each column has the same number of elements, and hence it is easy to perform CNN operations such as convolution and pooling. For hexagon-shaped tensors, different columns have different length and need to be considered separately.
	Based on the hexagonal data representation and storage in Fig.~\ref{fig:4}, HexCNN modifies the fundamental operations of CNN, including both forward and backward propagation algorithms~\cite{hecht1992theory,krizhevsky2012imagenet,lecun2015deep}. 
	
	\textbf{Forward-propagation}: Forward-propagation mainly contains convolution, pooling, and fully connection operations. 
	After flattening the output of convolution and pooling layers, the fully connected layer works the same as that for traditional rectangle-based CNN models. The convolution of HexCNN is represented as Equation~\ref{eq:conv},
	\begin{equation} 
	\setlength{\abovedisplayskip}{3pt}
	\setlength{\belowdisplayskip}{3pt}
	\begin{aligned}
	O_{u,v} = &\sum_{j=v}^{v+2\cdot l_k-1}\sum_{i=i_s}^{i_e}I_{i,j} \cdot k_{i-i_s,j-v}\\
	i_{s} =& \left\{
	\begin{aligned}
	j-l_I +1 + u& , & j \geq l_I \\
	u & , & j < l_I
	\end{aligned}
	\right.\\
	i_{e} =& \left\{
	\begin{aligned}
	i_s + 2\cdot l_k -1 & , & j \geq v+l_k \\
	i_s+l_k+(j-v) & , & j < v+l_k
	\end{aligned}
	\right.
	\end{aligned}
	\label{eq:conv}
	\end{equation}
	where input $I$ and filter $k$ are two-dimensional tensors in this example. $l_k$ and $l_I$ denote the side length of the filter and the input, respectively. $i_s$ and $i_e$ denote the start and the end indexes for each column in the sliding window, which are $u$ and $v$ dependant. $i_s$ increases with a step length of one after the first $l_I$ columns, and the range for index $j$ increases with a step length of one for the first $l_k$ columns and then decreases for the onward columns.
	
	For max-pooling and average-pooling, we either pick the maximum or average the value of all elements within a sliding window to obtain an output element.
	Take maxpooling as example, we iteratively fetch the elements from the red bounded area (output) in Fig.~\ref{fig:4}a and obtain the corresponding sliding window from the input. More formally, the procedure is represented as:
	\begin{equation} 
	\setlength{\abovedisplayskip}{3pt}
	\setlength{\belowdisplayskip}{3pt}
	O_{u,v} = \max\limits_{v\le j < v+2\cdot l_k-1,\, i_s\le i < i_e} I_{i,j}
	\label{eq:pool}
	\end{equation}
	\textbf{Backpropagation:} In backpropagation, the operations for dense layers are the same as those for rectangle-based frameworks.  For convolutional and pooling layers, we backpropagate the error and update the weights of each layer. 
	
	For back propagating the error in convolutional layers, HexCNN upsamples the error $\delta^{c}$ to $\hat{\delta^{c}}$ to get the same size as the input of the current layer $I$. HexCNN then performs full convolution on the upsampled $\hat{\delta^{c}}$ and the filter $k$ that has been rotated for 180 degrees. After the full convolution, the former layer's error $\delta^{p}$ is the Hadamard product~\cite{horn1990hadamard} with the derivative of the former layer's activation function $\sigma^{p}$:
	\begin{equation} 
	\setlength{\abovedisplayskip}{3pt}
	\setlength{\belowdisplayskip}{3pt}
	\delta^{p} = \hat{\delta}^{c} * rot180(k) \odot \sigma^{'p}
	\label{eq:bpc}
	\end{equation}
	where function $rot180$ rotates a matrix by 180 degrees, operator $*$ denotes the full convolution, and operator $\odot$ computes the Hadamard product.
	For pooling layers, we take max-pooling as an example to illustrate how HexCNN works. HexCNN first upsamples $\delta^{c}$ to $\hat{\delta^{c}}$, and figures out which position in $I$ corresponds to a specific value in $\delta^{c}$ (the perceptive field of $\delta^{c}$). HexCNN then transfers the value in $\delta^{c}$ to that position in $\hat{\delta}^{c}$, and sets the rest positions in $\hat{\delta}^{c}$ as zeros. As shown in Equation~\ref{eq:bpm}, each value in the former layer's error $\delta_{u,v}^{p}$ is the element-wise product of $\hat{\delta}_{u,v}^{c}$ and the sum of those values in $\delta^{c}$ that come from $I_{u,v}$.
	\begin{equation} 
	\setlength{\abovedisplayskip}{3pt}
	\setlength{\belowdisplayskip}{3pt}
	\delta_{u,v}^{p} = \hat{\delta}_{u,v}^{c} \cdot \sum_{I_{u,v} \rightarrow\delta_{i,j}} \delta_{i,j}
	\label{eq:bpm}
	\end{equation}
	where $I_{u,v} \rightarrow\delta_{i,j}$ denotes elements in $I$ that are within the perceptive filed of $\delta_{i,j}$, ``$\cdot$" computes the element-wise product.
	
	For updating the weights, we propose derivative computation algorithms for native hexagonal processing. HexCNN computes the partial derivative for both input and filters in convolutional layers, except the initial input of the network. The partial derivative of filters is for updating the weights, and the partial derivative of the input is for updating the weights of the former layer. \textbf{(i)} The partial derivative for a filter $k$ can be represented as the convolution between the input $I$ and the error $\delta$ (with the same size as the output $O$) according to the chain rule. We assume that ``VALID'' padding is applied. Then the partial derivative for the filter k can be represented as:
	\begin{equation} 
	\setlength{\abovedisplayskip}{3pt}
	\setlength{\belowdisplayskip}{3pt}
	\begin{aligned}
	\frac{\partial E}{\partial k_{u,v}} =& \sum_{j=v}^{v+2\cdot l_O-1}\sum_{i=i_{s}}^{i_{e}} I_{i,j}\cdot \delta_{i-i_{s},j-v}\\
	\end{aligned}
	\label{eq:bpk}
	\end{equation}
	where $i_{s}$ and $i_{e}$ are computed similar to Equation~\ref{eq:conv}.
	\textbf{(ii)} Computing the partial derivative of the input can be transformed into the full convolution between the error $\delta$ (with the same size of the output $O$) and the filter $k$. If the output size is smaller than the input size, HexCNN pads zeros around the $\delta$ to achieve the same size as the input before the full convolution. We summarize the procedure as:
	\begin{equation} 
	\setlength{\abovedisplayskip}{3pt}
	\setlength{\belowdisplayskip}{3pt}
	\begin{aligned}
	\frac{\partial E}{\partial I_{u,v}} =& \sum_{j=v-l_k+1}^{v+l_k}\sum_{i=i_s}^{i_e} \delta_{i,j}\cdot k_{i-i_s,j-v+l_k-1}\\
	i_{s} =& \left\{
	\begin{aligned}
	j -l_O  + u-l_k+2& , & j \geq l_O \\
	u-l_k+1 & , & j < l_O
	\end{aligned}
	\right.\\
	i_{e} =& \left\{
	\begin{aligned}
	i_s + 2\cdot l_k -1 & , & j \geq v \\
	i_s+2\cdot l_k+j-v -1& , & j < v
	\end{aligned}
	\right.
	\end{aligned}
	\label{eq:bp1}
	\end{equation}
	where $i_s$ and $i_e$ are column index $i$ and filter size $l_k$ related. $\delta_{i,j}$ returns zero when $i_s$ and $i_e$ out of the range of $\delta$ and this is the reason that we pad zeros around $\delta$.

	\begin{figure}[t]
		\centering
		\setlength{\belowcaptionskip}{-0.5cm}   
		\includegraphics[width=0.7\columnwidth]{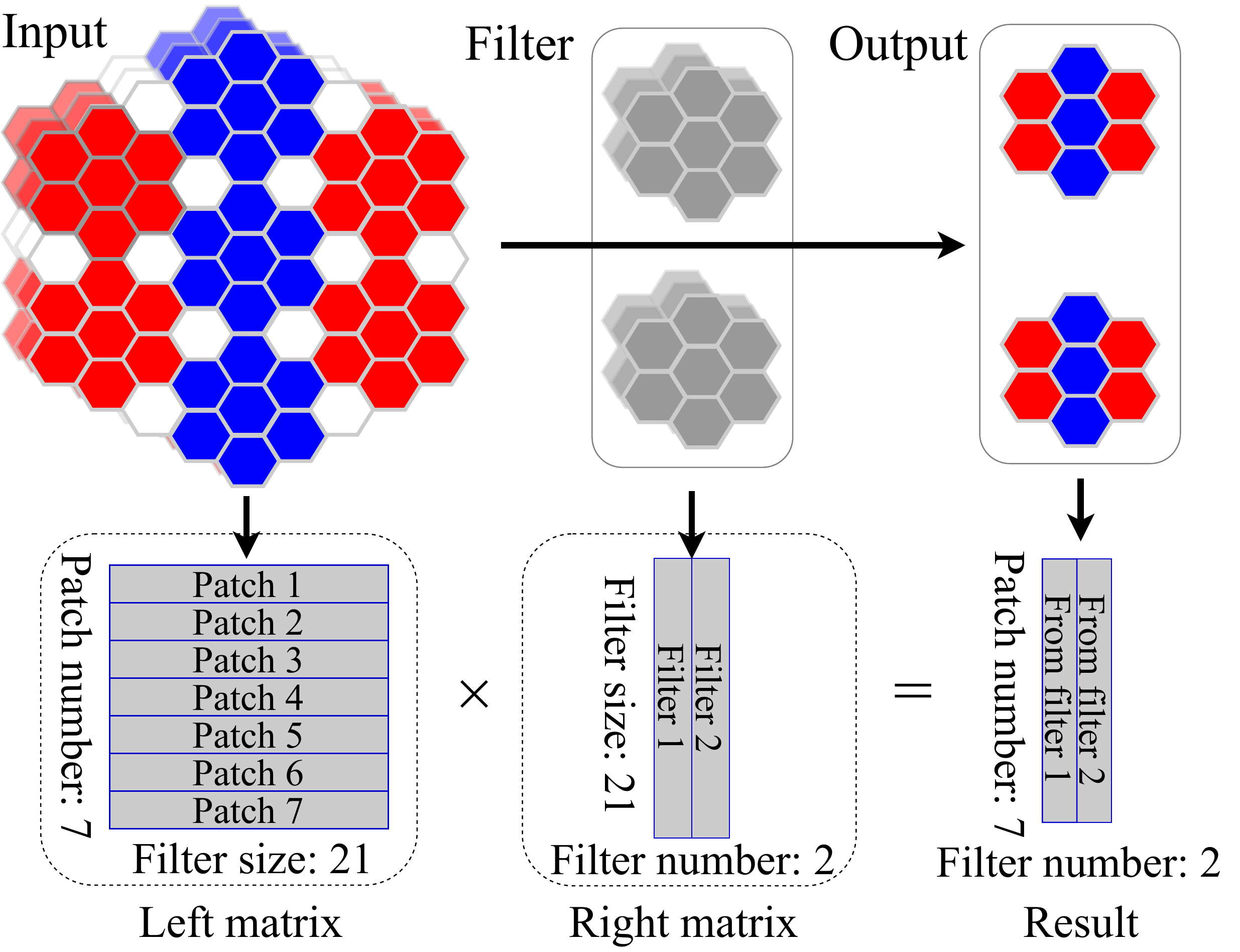}
		\caption{Hexagonal convolution to matrix multiplication. The red and blue color are used for locating differentiating different filters (sliding windows) only (best view in color).}
		\label{fig:5}
	\end{figure}
	
	
	\section{Transformation from Hexagonal Convolution to Matrix Multiplication}
	\label{sec:ctom}
	Existing deep learning frameworks transform convolution into matrix multiplication to speed up the computation. In this section, we present how HexCNN transforms hexagonal convolution into matrix multiplication. HexCNN splits the input into multiple patches according to the filter size and stride.
	Suppose the hexagon-shaped input has a side length $l_I$ and a channel $c$. The hexagonal convolution filters have a side length $l_k$, and the stride during convolution is $s$. HexCNN computes the number of patches by $l_p \cdot (l_p-1) + 1$, where $l_p = (l_I-l_k)/ s + 1$.
	Take Fig.~\ref{fig:5} as an example, where $l_I=5$, $c=3$, $l_k=2$, and $s=3$. The value of $l_p$ is two, and hence there are seven patches in the input (the red and blue colors are used for differentiation only). 
	
	To transform the hexagonal convolution into matrix multiplication, HexCNN first obtains the number of patches according to the input shape, the stride, and padding strategy. It then re-orders the elements in each patch as a vector in column-major to obtain the left matrix in Fig.~\ref{fig:5}. For the right matrix in Fig.~\ref{fig:5}, the width is the number of filters, and the height is the filter size. After transforming the input and filters into the left and right matrices, HexCNN obtains the convolutional output via matrix multiplication. To speed up the matrix multiplication, HexCNN then applies the GotoBLAS library~\cite{goto2008anatomy} for acceleration.
	
	\section{Experiments}
	\label{sec:experiment}
	Hexagon-imitation models have shown better accuracy than traditional rectangular CNN models on various applications and data sets~\cite{hoogeboom2018hexaconv,ke2018hexagon,luo2019hexagonal,sun2016design}. HexCNN has the same output as hexagon-imitation models, and we focus on the time and space efficiency of HexCNN in this section. 
	We first evaluate the time efficiency of HexCNN and then evaluate the space efficiency.
	We implement HexCNN by adapting the third-party Eigen library for the TensorFlow CPU version.
	We are unable to test the TensorFlow GPU version because the GPU version is based on cuDNN, which is a closed source library (same for Caffe and PyTorch).
	
	\textbf{Experimental setup:}
	\label{sec:parameters}
	The PC that we used is equipped with Intel(R) Core (TM) i7-7700 @ 3.60 GHz $\times$ 8 with 32GB memory.
	We implement two hexagon-imitation models based on TensorFlow. The first is the widely used ZeroOut~\cite{hoogeboom2018hexaconv}, and the second is the Quasi-H~\cite{sun2016design}.
	Note that Quasi-H simulates hexagonal processing on deep CNN models, and it is not hexagonal processing in the strict sense (see Section~\ref{hexForimg} for more details). We consider it as one of the baselines because it simulates hexagonal processing in certain circumstances.
	
	\begin{figure}[tbh]
		\centering
		\setlength{\belowcaptionskip}{-0.5cm}   
		\begin{subfigure}[a]{0.43\linewidth}
			\setlength{\abovecaptionskip}{0pt}   
			\includegraphics[width=\linewidth]{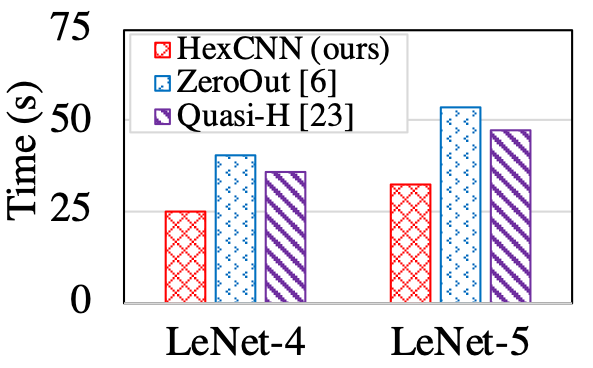}
			\caption{}
		\end{subfigure}\,
		\begin{subfigure}[a]{0.43\linewidth}
			\setlength{\abovecaptionskip}{0pt}   
			\includegraphics[width=\linewidth]{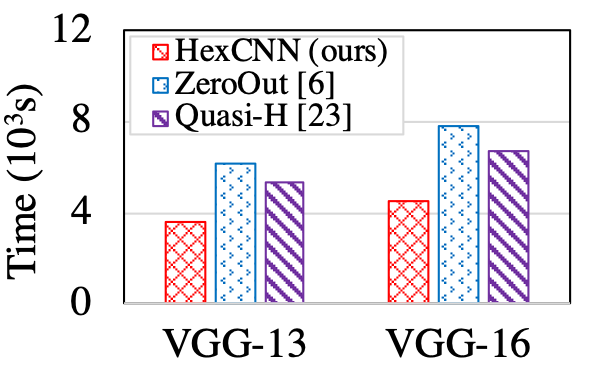}
			\caption{}
		\end{subfigure}
		\caption{Time consumption of HexCNN, ZeroOut and Quasi-H on (a) shallow and (b) deep hexagonal CNN models (best view in color).}
		\label{fig:6}
	\end{figure}
	
	\subsection{Time Efficiency Evaluation}
	\label{sec:effiHex}
	To evaluate the time efficiency of HexCNN, we compare the training time of a single batch (batch size as 40) in LeNet-4~\cite{lecun1998gradient}, LeNet-5~\cite{lecun1998gradient}, VGG-13, and VGG-16~\cite{simonyan2014very}. We keep the structure of these CNN models but perform hexagonal processing. 
	The synthetic data set that we used contains 2,000 hexagon-shaped input with a channel of three and a side length of 256 (511$\times$511 for ZeroOut, 511$\times$443 Quasi-H to keep the equivalent information). We obtain the data set from MNIST~\cite{lecun1998gradient} by resizing images into a size of 511$\times$511 followed by resampling them to hexagonal grids. We then crop them to the size needed.

	Fig.~\ref{fig:6}a shows the time consumption of different methods on two shallow hexagonal CNN models LeNet-4 and LeNet-5. HexCNN saves the training time by 38.7\% and 39.7\% compared with ZeroOut, respectively. Comparing with Quasi-H, HexCNN saves the training time by 30.8\% and 31.4\%, respectively.
	For deep hexagonal CNN models VGG-13 and VGG-16, HexCNN saves the training time by 41.3\% and 42.2\% compared with ZeroOut, respectively. In comparison with Quasi-H, HexCNN saves the training time by 32.6\% and 33.2\%, respectively.
	We find that on the same network structure, the more layers the models have, the more time that HexCNN saves. This is because the start-up cost is relatively stable, and the more layers, the more relative time that HexCNN saves.
	
	\subsection{Space Efficiency Evaluation}
	\label{sec:space}
	We compute the memory consumption of HexCNN, HexagDLy, ZeroOut, and Quasi-H for loading the input and the matrices for performing convolution to evaluate the space efficiency of HexCNN. We set the input channel as three, and the side length $x$ varies from 30 to 120. In convolution, we set the filter number as one, the side length of hexagon-shaped filters as two and keep the stride as one. 
	
	When loading the input, \textbf{(i)} ZeroOut method pads the input into a parallelogram with the side length of $2x-1$ to keep the equivalent information. Fig.~\ref{fig:1}b illustrates the input after padding, where the side length of the parallelogram is twice as that for the hexagonal input in Fig.~\ref{fig:1}a.
	\textbf{(ii)} Quasi-H re-samples and pads the input into a rectangle with a size of $2x-1$,$\sqrt{3}x$ to keep the equivalent information, as shown in Fig.~\ref{fig:1}c. 
	\textbf{(iii)} HexagDLy first rotates the hexagonal array in Cartesian coordinates to achieve a vertical alignment of neighboring elements and then aligns horizontally by shifting every second column upwards by half the distance between neighboring pixels.
	HexagDLy does not incur resampling because it takes hexagonal grids directly, and the input becomes a rectangle-shaped hexagonal grid with a size of $2x-1$ under the offset coordinate system.
	In convolution, ZeroOut and Quasi-H have a filter size of three to keep the equivalent information as HexCNN. 
	For HexagDLy, it divides the hexagon-shaped filters into multiple rectangle-shaped filters with different sizes, as shown in Fig.~\ref{fig:2}c. Therefore, the space for filters in HexagDLy is the same as that for HexCNN.
	
	\begin{figure}[tb]
		\centering
		\setlength{\belowcaptionskip}{-0.5cm}   
		\begin{subfigure}[a]{0.45\linewidth}
			\setlength{\abovecaptionskip}{0pt}   
			\includegraphics[width=\linewidth]{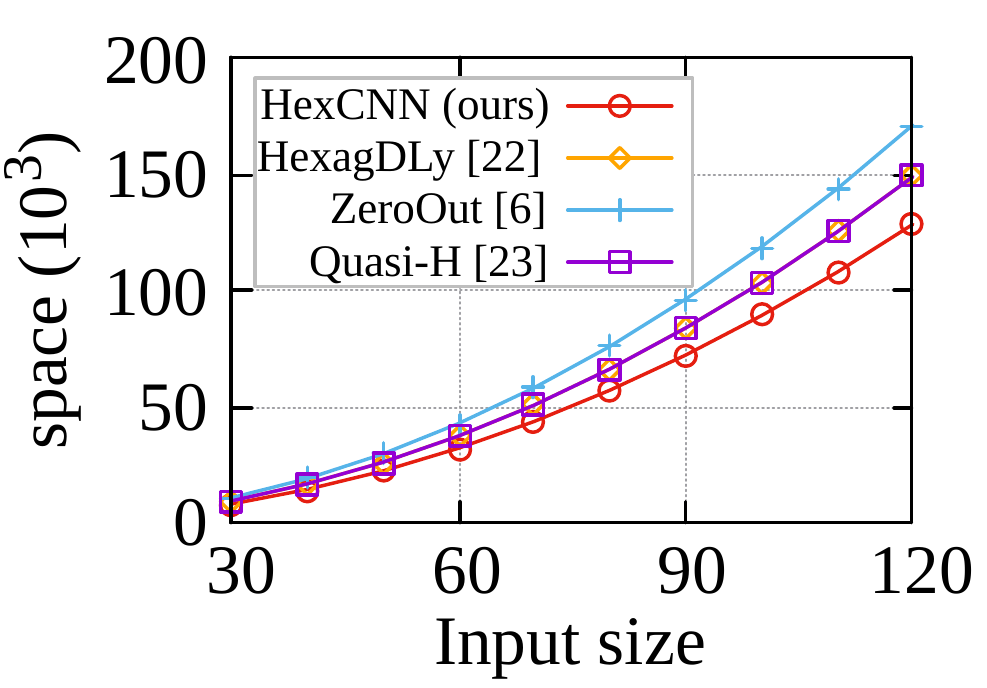}
			\caption{Space for input.}
		\end{subfigure}\,
		\begin{subfigure}[a]{0.45\linewidth}
			\setlength{\abovecaptionskip}{0pt}   
			\includegraphics[width=\linewidth]{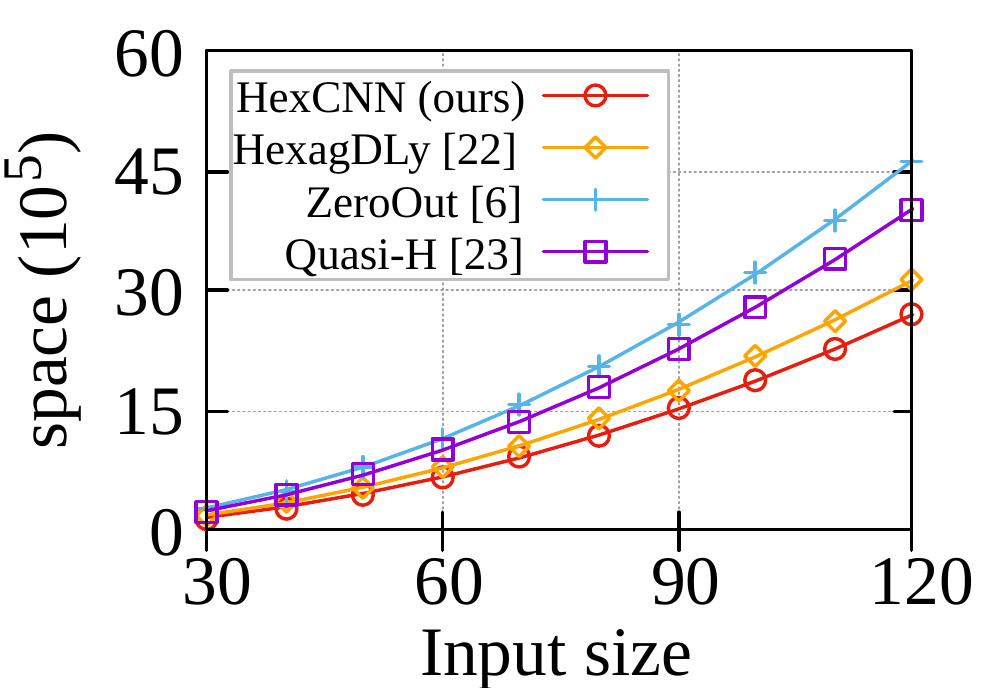}
			\caption{Space for convolution.}
		\end{subfigure}
		\caption{The space required by HexCNN, HexagDLy, ZeroOut, and Quasi-H for (a) loading one input and (b) preparing the corresponding matrices for convolution (best view in color).}
		\label{fig:8}
	\end{figure}
	
	As shown in Fig.~\ref{fig:8}, HexCNN saves the space cost significantly compared with existing hexagon-imitation models. With the increase of input size, HexCNN reduces the space consumption for storing input and performing convolution by 13.8\% compared with HexagDLy. Compared with ZeroOut, HexCNN reduces the space consumption for storing input and performing convolution by 25\% and 41.7\%, respectively. Compared with Quasi-H, HexCNN reduces the space consumption for storing the input and performing convolution by 13.8\% and 32.9\%, respectively.
	
	\subsection{HexCNN for Rectangle-Shaped Input}
	\label{sec:hexforrec}
	HexCNN achieves superior efficiency for hexagon-shaped input due to the seamless representation. 
	For rectangle-shaped input, HexCNN incurs resampling and padding. Fig.~\ref{fig:1}h shows the padding cost in HexCNN for square input. Suppose the side length of a square input is $x$, the side length of the corresponding hexagon that exactly covers the square input is $y$. Then $y$ has the following restriction: $y\geq \lceil(3x+1)/4\rceil$, and the padded area when $x$ is approaching infinity is 0.563$x^2$. 
	
	For ZeroOut, the padded area is 0.577$x^2$, as illustrated in Fig.~\ref{fig:1}f. Although the extra areas incurred by HexCNN and ZeroOut are similar, HexCNN still outperforms ZeroOut on the training time with both shallow and deep CNN models, because the averaged training time for each element is reduced, as illustrated in Section~\ref{sec:effiHex}. 
	Meanwhile, HexCNN achieves superior space efficiency in convolution compared with ZeroOut, as illustrated in Section~\ref{sec:space}. 
	
	Quasi-H takes rectangle-shaped input directly without padding, as shown in Fig.~\ref{fig:1}g. In convolution, Quasi incurs unnecessary padding for the filters, so as to apply existing deep learning frameworks.
	Although the padded area for the filters in convolution leads to deficiency, the overall efficiency of Quasi-H on rectangle-shaped input is still superior to HexCNN.
	However, Quasi-H simulates hexagonal processing on deep CNN models. It is not hexagonal processing in the strict sense (see Section~\ref{hexForimg} for more details).
	
	\section{Conclusions and future work}
	\label{sec:conclusion}
	We proposed a native hexagonal CNN framework named HexCNN. 
	HexCNN directly takes hexagon-shaped input and performs native hexagonal processing for both forward and backward propagation based on hexagon-shaped filters. It transforms hexagonal convolution to matrix multiplication to achieve high time efficiency. Experimental results on complete CNN models show that HexCNN outperforms hexagon-imitation models by reducing up to 42.2\% of the training time. Meanwhile, HexCNN saves the space by up to 25\% for loading the input, and up to 41.7\% for performing convolution. 
	HexCNN shows great performance from our experimental results on CPU. In the future, we will implement complete forward and backward propagation of HexCNN on GPU to obtain higher applicability of the framework.

	\bibliographystyle{plain}
	\bibliography{ICDM}
	\balance
	
\end{document}